\documentclass[10pt,twocolumn,letterpaper]{article}

\usepackage{cvpr}              

\usepackage{booktabs}
\usepackage{tabularx,array}
\usepackage{threeparttable}
\usepackage{capt-of}
\usepackage[table]{xcolor}
\definecolor{heatY}{RGB}{255,247,173}  
\definecolor{heatT}{RGB}{255,231,199}  
\definecolor{heatR}{RGB}{255,220,220}  
\newcommand{\Y}[1]{\cellcolor{heatY}#1}
\newcommand{\T}[1]{\cellcolor{heatT}#1}
\newcommand{\R}[1]{\cellcolor{heatR}#1}
\newcolumntype{G}{@{\hspace{0pt}}p{12pt}@{\hspace{0pt}}}
\newcolumntype{Y}{>{\centering\arraybackslash}X}

\definecolor{cvprblue}{rgb}{0.21,0.49,0.74}
\usepackage[pagebackref,breaklinks,colorlinks,allcolors=cvprblue]{hyperref}

\def\paperID{} 
\def\confName{CVPR}
\def\confYear{2026}

\title{Advancing Structured Priors for Sparse-Voxel Surface Reconstruction}

\author{
Ting-Hsun Chi$^{1}$ \and
Chu-Rong Chen$^{1}$ \and
Chi-Tun Hsu$^{1}$ \and
Hsuan-Ting Lin$^{1}$ \and
Sheng-Yu Huang$^{1,2}$ \and
Cheng Sun$^{2}$ \and
Yu-Chiang Frank Wang$^{1,2}$ \and
\makebox[\linewidth]{\centering
$^{1}$National Taiwan University \quad
$^{2}$NVIDIA
}
}

\begin{document}
\twocolumn[{%
\maketitle
\begin{center}
  \includegraphics[width=\textwidth]{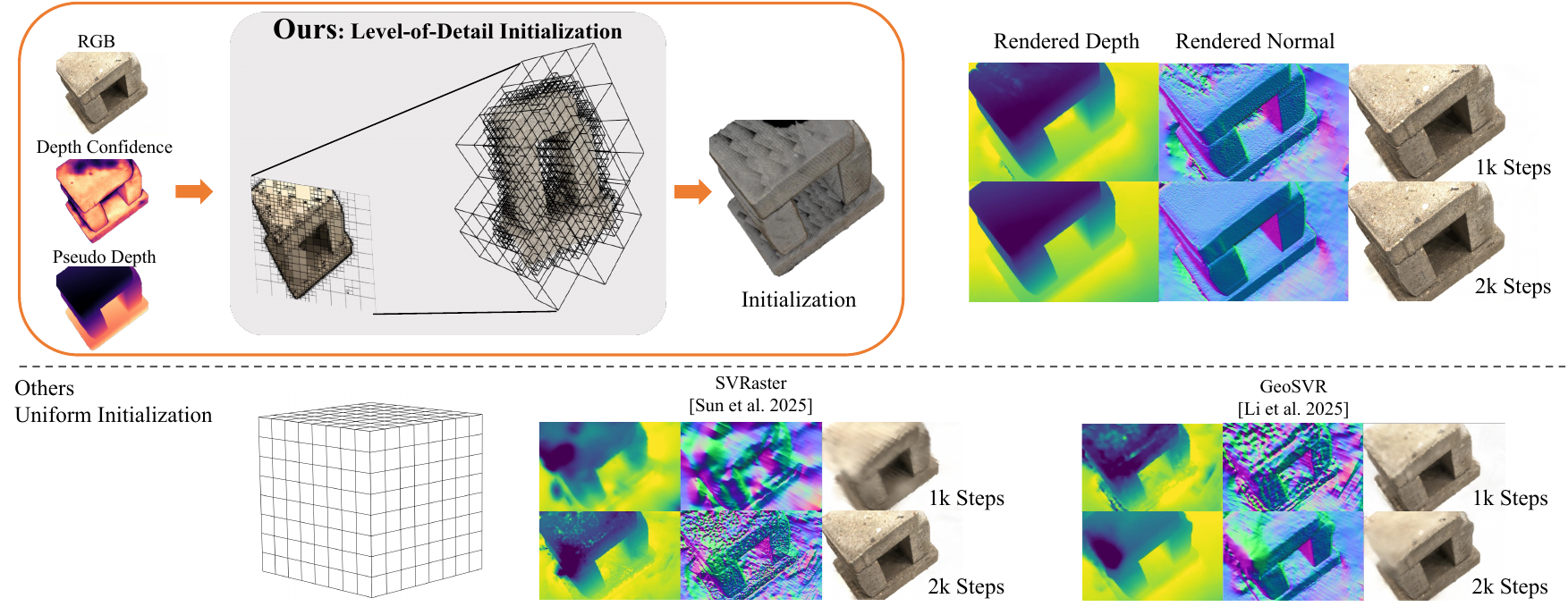}
  \captionof{figure}{Our method initializes voxels at plausible regions with the associated levels of detail. With the geometry is refined by direct depth supervision, our SVRaster reconstruction better captures detailed 3D geometry. Our initialization-seeded design couples fast convergence with high geometric fidelity and strong completeness throughout training.
}
  \label{fig:teaser}
\end{center}
}]

\begin{abstract}
Reconstructing accurate surfaces with radiance fields has progressed rapidly, yet two promising explicit representations, 3D Gaussian Splatting and sparse-voxel rasterization, exhibit complementary strengths and weaknesses. 3D Gaussian Splatting converges quickly and carries useful geometric priors, but surface fidelity is limited by its point-like parameterization. Sparse-voxel rasterization provides continuous opacity fields and crisp geometry, but its typical uniform dense-grid initialization slows convergence and underutilizes scene structure.
We combine the advantages of both by introducing a voxel initialization method that places voxels at plausible locations and with appropriate levels of detail, yielding a strong starting point for per-scene optimization. To further enhance depth consistency without blurring edges, we propose refined depth geometry supervision that converts multi-view cues into direct per-ray depth regularization.
Experiments on standard benchmarks demonstrate improvements over prior methods in geometric accuracy, better fine-structure recovery, and more complete surfaces, while maintaining fast convergence.
\end{abstract}    
\section{Introduction}
\label{sec:intro}
Surface reconstruction from multi-view images has long been a fundamental problem in computer vision, with broad applications in robotics, gaming, and AR/VR. Despite extensive prior research, achieving both fast and accurate surface reconstruction remains an ongoing challenge.

The advent of Neural Radiance Fields (NeRF) ~\cite{mildenhall2021nerf} has brought significant progress to novel view rendering and surface reconstruction by learning continuous implicit representations through differentiable volumetric rendering ~\cite{yariv2021volsdf, wang2021neus,wang2024neurodin,li2023neuralangelo}. However, most NeRF-based methods are computationally expensive, limiting their practicality in high-resolution or large-scale scenarios. Recently, 3D Gaussian Splatting (3DGS)~\cite{kerbl20233d} has emerged as a promising alternative that leverages explicit Gaussian primitives with differentiable rasterization to achieve real-time, high-quality rendering. Its remarkable performance in novel view synthesis has encouraged extensions toward surface reconstruction~\cite{guedon2024sugar, huang20242d, yu2024gaussian, chen2024pgsr}. Nevertheless, 3DGS still faces inherent limitations: it depends heavily on multi-view point cloud initialization, which is often incomplete or noisy, and its soft Gaussian formulation lacks explicit surface boundaries, leading to geometric ambiguity.

Motivated by the limitations of 3DGS, recent work has revisited sparse voxel-based representations. SVRaster ~\cite{sun2025sparse} replaces Gaussian primitives with a \textit{sparse voxel octree} (SVO) and employs rasterization for real-time rendering, mitigating the geometric ambiguity inherent in 3D Gaussian primitives while addressing the inefficiency of traditional voxel-based methods. Despite the advances in renderable representations ~\cite{mildenhall2021nerf, kerbl20233d,sun2025sparse}, e.g. NeRF, 3DGS, and sparse voxels, surface reconstruction still suffers from the lack of explicit geometry constraints, which often hinders stable optimization and overall reconstruction accuracy. 

To address this bottleneck, recent studies incorporate external priors such as pretrained monocular depth maps, geometry diffusion model, and 3D foundation model cues to regularize optimization. 
Numerous 3DGS-based frameworks~\cite{fan2024instantsplat, xu2025depthsplat, chen2024pgsr} adopt these priors for initialization or supervision to improve geometry. 
Sparse voxel-based approaches~\cite{sun2025sparse,ligeosvr} further show this structured representation, regularized with foundation model generated prior, outperforms 3DGS-based methods in geometry accuracy. 

Although sparse-voxel formulations offer a promising path toward structured and prior-guided reconstruction, existing approaches largely rely on uniform initialization, resulting in inefficient voxel allocation, slow convergence, and limited exploitation of geometric priors. 


In this paper, we propose a sparse voxel reconstruction framework with a Voxel Initialization with Level of Details algorithm that directly places geometrically valid voxels with adaptive, detail-aware resolution guided by multi-view geometry cues. 
To handle uncertainty in pseudo geometry priors, we introduce an uncertainty-aware opacity field that stabilizes incomplete regions and preserves surface continuity. To avoid overly smooth surfaces, we regularize the geometry with Refined Depth for Geometry Supervision in per-scene optimization. 


\noindent The key contributions of this work are as follows :

\begin{itemize}
     \item We design a geometry-guided voxel placement strategy that infers both position and resolution  from multi-view foundation model cues. This adaptive  initialization leverages scene structure to form a compact yet accurate voxel hierarchy, reducing reliance on dense-grid initialization.

    \item We develop a topology-aligned multi-view fusion mechanism that aligns per-view octrees into a globally consistent structure, integrating uncertain or incomplete priors through an uncertainty-aware opacity formulation.

    \item We propose a lightweight depth-regularization objective that refines the reconstructed geometry during per-scene training, improving local sharpness and depth consistency without sacrificing efficiency.

    \item Our method yields more geometrically faithful surfaces with fast convergence compared to existing methods, bridging efficient initialization and high-fidelity surface reconstruction.
\end{itemize}

\section{Related Work}
\label{sec:related_work}

\subsection{Radiance Fields and Neural Rendering}

A radiance field refers to a function that maps every point in 3D space to its emitted radiance and volume density. Neural Radiance Fields (NeRF)~\cite{mildenhall2021nerf} pioneered the use of neural implicit radiance field representations for high-quality 3D reconstruction and novel-view synthesis by encoding scene attributes with an MLP and rendering via differentiable volume integration. Subsequent work improves NeRF by introducing anti-aliased and multiscale scene representations~\cite{barron2021mip, barron2023zip, hu2023tri}, and by greatly accelerating training and rendering through multiresolution hash-encoded features~\cite{muller2022instant}. However, these methods still rely on implicit neural fields that require dense per-ray sampling and continuous MLP evaluation, making them slow to render, difficult to edit, and inefficient to scale.

3D Gaussian Splatting (3DGS)~\cite{kerbl20233d} represents scenes using explicit anisotropic Gaussian primitives and renders images via efficient differentiable rasterization, enabling high-quality reconstruction with real-time performance. However, the reliance on depth sorting and the overlap of Gaussians may induce several view-inconsistent rendering problems. Although recent methods attempt to mitigate these issues~\cite{radl2024stopthepop, moenne20243d}, representations with more stable volumetric definitions and inherent ordering remain desirable.

Sparse voxel structures, which provide well-defined spatial occupancy and natural spatial ordering, have already been widely used to represent 3D scenes.
While some prior works use sparse voxels combined with MLPs to decode radiance or surface features locally~\cite{liu2020neural, wu2022voxurf, li2022vox}, others adopt more explicit voxel representations~\cite{sun2022direct, fridovich2022plenoxels} to improve training speed. SVRaster~\cite{sun2025sparse} integrates sparse voxels with the fast rasterization strategy of 3DGS. By adaptively adjusting voxel sizes to match scene complexity, SVRaster achieves a balance among memory efficiency, rendering quality, and runtime performance.

\subsection{Learnable Fields for Surface Reconstruction}

Surface reconstruction from multi-view images has been an important task in computer vision. Classical multi-view stereo (MVS) methods reconstruct surfaces from calibrated images via modular pipelines of structure-from-motion, depth estimation, and fusion. PatchMatch-based methods~\cite{furukawa2009accurate} and COLMAP~\cite{schonberger2016pixelwise} enabled efficient, high-quality reconstructions. As deep learning entered the field,  MVSNet~\cite{yao2018mvsnet} improved matching through deep cost volumes but retained the multi-stage design. 

Neural implicit methods introduced a new paradigm by representing surfaces as continuous fields using MLPs. Early works like DeepSDF~\cite{park2019deepsdf} and IDR~\cite{yariv2020multiview} demonstrated that signed distance functions could be learned directly from multi-view images via differentiable rendering. Later, methods such as UNISURF~\cite{oechsle2021unisurf}, NeuS ~\cite{wang2021neus}, and VolSDF~\cite{yariv2021volume} integrated volume rendering with implicit surfaces to achieve sharper and more accurate reconstructions. However, these implicit representations often require expensive optimization, making them slow to train on complex scenes.

Following the introduction of 3D Gaussian Splatting~\cite{kerbl20233d}, a new line of surface reconstruction methods emerged based on Gaussian primitives. SuGaR~\cite{guedon2024sugar} aligns Gaussians with mesh geometry under regularization. Later works integrated neural SDFs or implicit fields to improve surface fidelity~\cite{chen2023neusg, yu2024gsdf}, and alternative formulations such as surfel-based representations were proposed to improve alignment and geometric sharpness~\cite{huang20242d, dai2024high}. More recently, several methods have incorporated 3D foundation models~\cite{wang2025vggt, wang2024dust3r, leroy2024grounding, yang2024depth} to guide reconstruction to better results. For example, InstantSplat~\cite{fan2024instantsplat} uses DUSt3R to initialize Gaussians from dense point maps, while Splatt3R~\cite{smart2024splatt3r} employs MASt3R~\cite{leroy2024grounding} to infer Gaussian parameters directly from two unposed images. DepthSplat~\cite{xu2025depthsplat} leverages Depth Anything V2~\cite{yang2024depth} to supervise depth estimation. While these methods benefit from strong geometric priors, their performance remains constrained by the need for high-quality initialization~\cite{chen2024pgsr, chen2024vcr}, and by the lack of clearly defined surfaces and edges of 3D Gaussian primitives.

Voxels offer a more structured and explicit representation of geometry, making them inherently better suited for surface reconstruction. For instance, SVRaster~\cite{sun2025sparse} achieves comparable speed and performance on several 3D tasks to state-of-the-art Gaussian-based methods, while benefiting from a clearer geometric structure. GeoSVR~\cite{ligeosvr} takes a step in this direction by combining voxel-based neural fields with geometric supervision from Depth Anything V2~\cite{yang2024depth}. It employs a sparse voxel grid to encode the scene and uses foundation model predictions as regularization signals. While it demonstrates improved reconstruction quality, the progressive pruning and subdivision starting from uniform voxel grids lead to slow convergence.

In this work, we address this bottleneck by directly initializing sparse voxels at geometrically meaningful locations using multi-view depth cues, enabling faster convergence, higher fidelity reconstruction, and improved alignment with scene structure from the very beginning.


\begin{figure*}[t]
    \centering
    \includegraphics[width=0.95\linewidth]{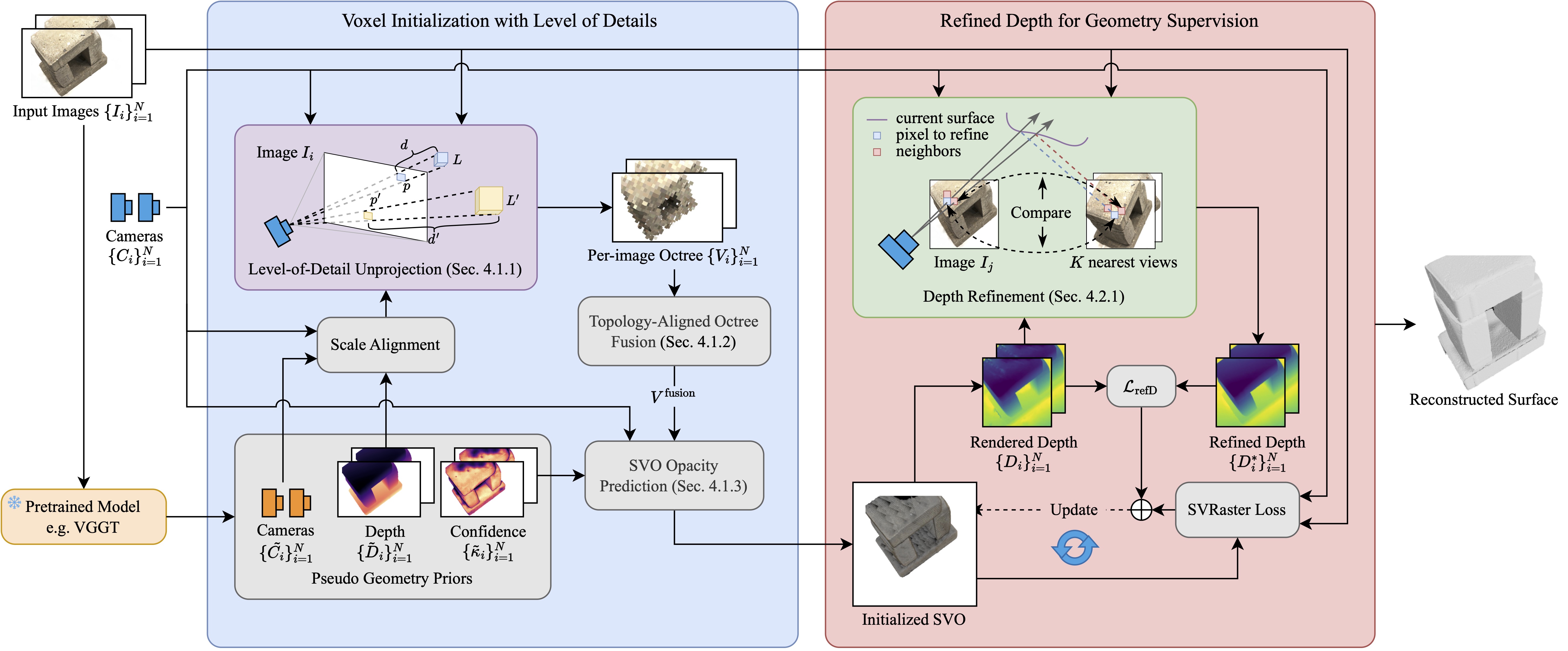}
    \caption{\textbf{Overview of Our Method.} Given images $\{I_i\}$ with cameras $\{C_i\}$, our method reconstructs surfaces via per-scene training. First, a pretrained model estimates pseudo geometry priors (i.e., cameras $\{\tilde{C_i}\}$, depth $\{\tilde{D_i}\}$, and confidence $\{\tilde{\kappa_i}\}$), which are utilized to build per-image octrees $\{V_i\}$ with our level-of-detail unprojection. The resulting per-image octrees are fused into a single octree $V^{\mathrm{fusion}}$ (as an initialized SVO), along with the prediction of the opacity of each fused voxel grid. In the per-scene SVO optimization stage, we derive the refined depths for each camera view via advancing the geometry supervision (i.e., $\mathcal{L}_{\mathrm{refD}}$) from the rendered depths $\{D_i\}$ and the associated input images. This geometry-supervision loss, when combined with the original SVRaster training loss, improves the geometric accuracy of the reconstructed surface.}
    \label{fig:overview}
\end{figure*}


\begin{figure}[t]
    \centering
    \includegraphics[width=0.95\linewidth]{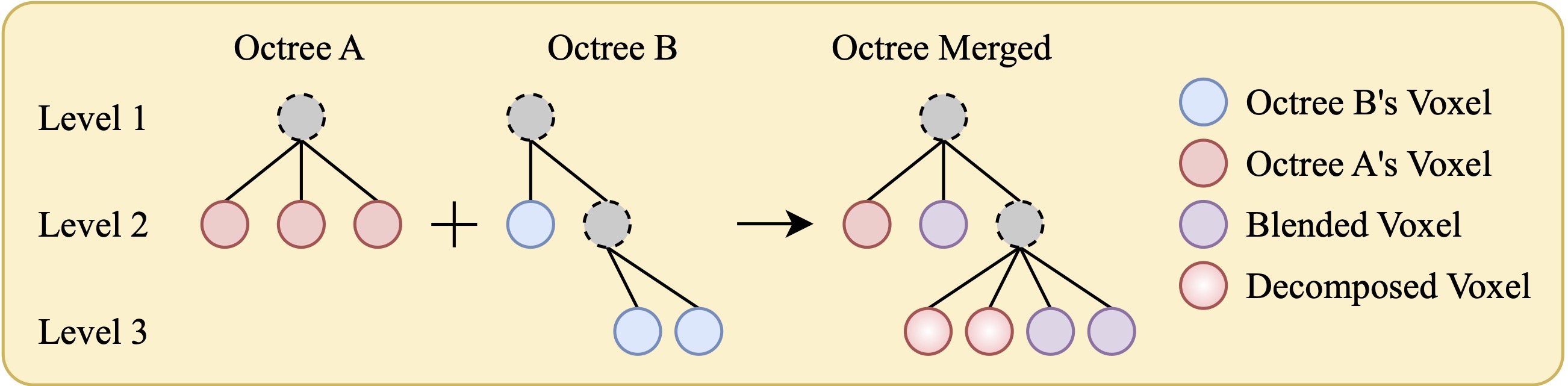}
    \caption{\textbf{Topology-Aligned Octree Fusion.} To integrate sparse octrees derived from different camera views, the final resolution of each voxel is determined by the per-view finest voxels occupying the same spatial cell. For example, the rightmost voxel of octree A at level 2 is decomposed into multiple level-3 voxels to align the structure of octree B. The decomposed voxels inherit properties (e.g. color) from their parent voxel. To obtain the final fused octree, we average the colors of the corresponding voxels across per-view octrees as the blended voxels (denoted by the purple voxels.}
    \label{fig:topo_align}
\end{figure}

\section{Preliminaries}
 \textbf{Sparse Voxel Rasterization}~\cite{sun2025sparse} models a scene as a sparse octree of voxels stored in a flat array. 
A voxel at level~$\ell$ has side length $s_\ell = S / 2^\ell$ and center
\begin{equation}
\begin{aligned}
\mathbf{c}_{i,j,k,\ell} &= \mathbf{c}_{\text{scene}} + s_\ell 
\left[ \mathbf{g}_{i, j, k} - 0.5 \!\cdot\! 2^\ell \!\cdot\!(1, 1, 1) \right],\\
\mathbf{g}_{i, j, k} &= (i{+}0.5,\, j{+}0.5,\, k{+}0.5).
\end{aligned}
\end{equation}
Each voxel stores eight corner densities $\{\sigma_m\}_{m=1}^8$ and a view-dependent color represented by spherical harmonics $f_{\text{SH}}(\mathbf{d})$. 
Inside a voxel, the density is trilinearly interpolated as
\begin{equation}
\sigma(\mathbf{u}) = \sum_{m=1}^{8} w_m(\mathbf{u}) \, \sigma_m, 
\qquad 
\rho(\mathbf{u}) = \mathrm{ELU}(\sigma(\mathbf{u})) + 1,
\end{equation}
where $\mathbf{u}\in[0,1]^3$ and $\rho$ is the non-negative density after activation.
The surface normal is analytically computed by
\begin{equation}
\mathbf{n} = 
\frac{\nabla_{\mathbf{x}} \rho(\mathbf{x})}
{\|\nabla_{\mathbf{x}} \rho(\mathbf{x})\|}.
\end{equation}

Rendering follows front-to-back alpha compositing:
\begin{equation}
\begin{aligned}
C      &= \sum_k T_k \, \alpha_k \, \mathbf{c}_k,\\
T_k    &= \prod_{j<k} (1 - \alpha_j),\\
\alpha_k &= 1 - \exp(-\rho_k \, \Delta t_k).
\end{aligned}
\end{equation}
where $\rho_k$ and $\Delta t_k$ denote the density and segment length of the $k$-th voxel along a ray.
Voxels are sorted using a \emph{direction-dependent Morton order} to ensure correct depth ordering across octree levels. 

During optimization, SVRaster alternates between 
\emph{pruning} voxels with negligible contribution $(T_k\alpha_k < \tau_p)$ 
and \emph{subdivision} of high-priority voxels. 
Parameters of new voxels are initialized by trilinear interpolation of corner densities and direct copying of spherical-harmonic coefficients. 
This adaptive, explicit field enables efficient differentiable rendering with coarse-to-fine geometric control.

\section{Method}
Fig.~\ref{fig:overview} provides an overview of our method, which reconstructs high-quality surfaces from a set of posed images using SVO in a scene-specific setting. The method consists of Voxel Initialization with Level of Detail (Sec.~\ref{sec:voxel_init}) that constructs an SVO for further per-scene optimization. In the per-scene optimization, the SVO geometry is refined with Refined Depth for Geometry Supervision (Sec.~\ref{sec:depth_loss}).

\subsection{Voxel Initialization with Level of Details }\label{sec:voxel_init}
First, we leverage the availability of multiple images covering the scene and employ a pretrained 3D vision foundation model to estimate pseudo geometric priors for each view. To accurately initialize voxel positions and sizes, we propose a level-of-detail unprojection algorithm and generate per-image SVOs (Sec.~\ref{sec:vox_unproject}). The per-image octrees are then merged into a single octree using our topology-aligned fusion algorithm (Sec.~\ref{sec:merge}). The opacity of each voxel grid point is assigned according to an uncertainty-aware opacity field (Sec.~\ref{sec:opacity}) to build an SVO as the initialization for per-scene optimization.

\subsubsection{Level-of-Detail Unprojection}\label{sec:vox_unproject}

Given $N$ calibrated images $\{I_i\}_{i=1}^{N}$ with camera parameters $\{C_i\}_{i=1}^{N}$, 
we construct corresponding SVOs $\{V_i\}_{i=1}^{N}$, 
where each $V_i$ represents a per-view initialization of the scene geometry. 
We adopt the foundation model VGGT~\cite{wang2025vggt} to obtain pseudo depths $\{\tilde{D}_i\}$, 
confidence maps $\{\kappa_i\}$, and estimated camera poses $\{\tilde{C}_i\}$, 
which are aligned to the real poses using Umeyama alignment~\cite{umeyama2002least}. 
The resulting similarity transform $(\mathbf{R}_i, \mathbf{t}_i, s)$ provides both rotation and translation alignment, 
and the global scale $s$ is also applied to $\tilde{D}_i$ for consistent depth scaling.
    
To unproject each pixel $p=(u,v)$ in $I_i$, we use the pseudo depth $\tilde{D}_i(u,v)$ 
and approximate its 3D footprint by sampling the pixel center and four half-offset neighbors:
\begin{equation}
\mathbf{x}_{(\cdot)} = 
\mathbf{R}_i^{\top}\!\left(
\tilde{D}_i(u',v')\,\mathbf{K}_i^{-1}[u',v',1]^{\top}-\mathbf{t}_i
\right),
\end{equation}
where $(u',v')\!\in\!\{(u,v),(u{\pm}0.5,v),(u,v{\pm}0.5)\}$ 
and $\mathbf{K}_i$ is the intrinsic matrix.
The spatial footprint of the pixel is approximated by
\begin{equation}
A_{i,p} \approx 
\big\|(\mathbf{x}_{u{+}0.5}-\mathbf{x}_{u{-}0.5}) 
\times (\mathbf{x}_{v{+}0.5}-\mathbf{x}_{v{-}0.5})\big\|.
\end{equation}

\textbf{Enforcing level awareness.}
A pixel covering a larger 3D footprint corresponds to lower local detail, 
since its color integrates information over a wider surface region. 
We therefore adapt the voxel resolution to the pixel’s effective spatial coverage.
The level $\ell_{i,p}$ of the voxel corresponding to pixel $p$ is selected as the largest level (smallest voxel size) whose voxel still covers the footprint,
\begin{equation}
\ell_{i,p} = \max_{\ell}\{\,\ell~|~s_\ell^{2}\ge A_{i,p}\,\},
\qquad
s_\ell = S/2^{\ell}.
\end{equation}
Pixels with large $A_{i,p}$ (low detail) are assigned coarser voxels, 
while those with small $A_{i,p}$ (high detail) receive finer voxels for accurate reconstruction.

Each pixel color $\mathbf{c}_{i,p}$ initializes a voxel $v_{i,p}$ at level~$\ell_{i,p}$ 
with zero-degree spherical harmonic coefficient $h^{0}_{i,p}$ following 3DGS~\cite{kerbl20233d}. 
We then check and merge all the voxels $\mathcal{V}_{v'}$ that can be included in a coarser voxel $v'$ at level~$\ell'$ 
if their color deviation
\begin{equation}
d_{v'} = 
\max_{v_k\in\mathcal{V}_{v'}}\|\mathbf{c}_k-\bar{\mathbf{c}}_{v'}\|_2,
\quad
\bar{\mathbf{c}}_{v'}=\frac{1}{|\mathcal{V}_{v'}|}\sum_{v_k\in\mathcal{V}_{v'}}\mathbf{c}_k,
\end{equation}
is below a threshold~$t$.
To prevent undersampling, we further enforce a minimum occupancy constraint:
\begin{equation}
\ell_{\min}(v')=\min_{v_k\in\mathcal{V}_{v'}}\ell_k,\quad
n_{\mathrm{req}}(v')=2^{\max(\ell_{\min}(v')-\ell',0)+1},
\end{equation}
and perform merging only if
\begin{equation}
d_{v'}<t
\quad\text{and}\quad
|\mathcal{V}_{v'}|\ge n_{\mathrm{req}}(v').
\end{equation}
This ensures that merges to much coarser levels occur only when enough fine voxels support the region. The merging process preserves geometric fidelity while reducing redundancy. We construct one SVO for each image, and all the octrees share the same world coordinate.

\subsubsection{Topology-Aligned Octree Fusion}\label{sec:merge}
Given the per-view octrees $\{V_i\}_{i=1}^{N}$ constructed in Sec.~\ref{sec:voxel_init}, 
we fuse them into a single global octree $V^{\mathrm{fusion}}$ that serves as the initialization for subsequent optimization. 
Each voxel $v \in V_i$ is parameterized by an octree level $\ell_{i,k}$ and a Morton-encoded path 
$p_{i,k}$ that uniquely identifies its spatial cell at that level. 
Our fusion procedure consists of two stages: a \emph{topology alignment} step that builds a common refinement of all octrees, followed by a \emph{feature aggregation} step that averages voxel attributes
across views. The fusion process is demonstrated in Fig.~\ref{fig:topo_align}.

\paragraph{Topology alignment.}
Different views may allocate voxels at different resolutions in the same spatial region. 
To obtain a consistent topology, we construct the finest common refinement of all per-view octree.
Let $L_{\min}$ and $L_{\max}$ denote the minimum and maximum octree levels appearing across all 
$\{V_i\}$. 
We iterate from level $L_{\min}$ to $L_{\max}-1$, and at each level $L$ we detect regions where some octrees are coarser than others.

Conceptually, for every voxel $v$ at level $L$ in any octree, we consider its Morton prefix at
level~$L$. 
If there exists another octree that already contains one or more voxels at level $\ell \ge L{+}1$
whose Morton prefixes coincide with that of $v$, the region covered by $v$ has been subdivided more finely in another view. 
In this case, we recursively subdivide $v$ into its eight children at level $L{+}1$, inheriting the parent's attributes to initialize the children. 
This procedure is applied symmetrically across all octrees and levels until no voxel has a
coarser representation than any other octree in the same spatial cell. 
The alignment ensures that all octrees share a compatible hierarchy and any occupied spatial cell is represented by voxels with identical $(p,\ell)$ across all views that cover it.

\paragraph{Feature aggregation.}
Once the topology is aligned, we fuse the voxel attributes across views by aggregating all voxels
that correspond to the same spatial cell. 
Specifically, we take the union of all voxels
\[
\mathcal{V}^{\mathrm{all}} = \bigcup_{i=1}^{N} V_i,
\]
and group them by their Morton path and level. 
For each cell $(p,\ell)$, let
\[
\mathcal{V}_{p,\ell} = \{\,v_{i,k} \in \mathcal{V}^{\mathrm{all}} 
\mid (p_{i,k}, \ell_{i,k}) = (p,\ell)\,\}
\]
denote the set of voxels that occupy this cell. 
Their zero-degree spherical-harmonic coefficients (colors) $\{h^{0}_{i,k}\}$ and higher-degree
coefficients $\{h^{\mathrm{sh}}_{i,k}\}$ are combined by averaging:
\begin{equation}
\bar{h}^{0}_{p,\ell} = \frac{1}{|\mathcal{V}_{p,\ell}|}
\sum_{v_{i,k} \in \mathcal{V}_{p,\ell}} h^{0}_{i,k},
\qquad
\bar{h}^{\mathrm{sh}}_{p,\ell} = \frac{1}{|\mathcal{V}_{p,\ell}|}
\sum_{v_{i,k} \in \mathcal{V}_{p,\ell}} h^{\mathrm{sh}}_{i,k}.
\end{equation}
The fused octree $V^{\mathrm{fusion}}$ is then defined by the set of unique cells $\{(p,\ell)\}$ and their aggregated attributes $\{\bar{h}^{0}_{p,\ell}, \bar{h}^{\mathrm{sh}}_{p,\ell}\}$. 
This multi-view fusion produces a globally consistent, multi-resolution octree that integrates per-view detail while preserving a compact representation.

\subsubsection{SVO Opacity Prediction}\label{sec:opacity}

While the fused octree $V^{\mathrm{fusion}}$ provides a globally consistent topology, 
its occupancy may still be incomplete or noisy due to inaccuracies in the estimated depth maps 
and view-dependent reconstruction gaps. 
To address these issues, we introduce an \emph{uncertainty-aware opacity assignment} stage 
that leverages truncated signed distance fusion (TSDF)~\cite{newcombe2011kinectfusion} weighted by per-pixel confidence maps.

\paragraph{TSDF-guided octree completion.}
We first combine the fused octree $V^{\mathrm{fusion}}$ with a uniform voxel grid of the same spatial extent, 
forming an augmented voxel grid $V^g$ that ensures all spatial regions within the scene bounds are represented. 
For each grid point $\mathbf{x}_j \in V^g$, we compute a signed distance value
\begin{equation}
    F(\mathbf{x}_j) = 
    \frac{\sum_{i} \kappa_i(\mathbf{x}_j)\, f_i(\mathbf{x}_j)}{\sum_{i} \kappa_i(\mathbf{x}_j)},
\end{equation}
where $f_i(\mathbf{x}_j)$ is the signed distance from $\mathbf{x}_j$ to the surface observed 
in view~$i$, and $\kappa_i(\mathbf{x}_j)$ denotes the pixel-wise confidence of the pseudo geometry priors. 
The signed distance values are truncated to $[-1,1]$ according to the truncation distance 
$d_{\mathrm{trunc}}$, following standard TSDF fusion~\cite{newcombe2011kinectfusion}. 
This produces an \emph{uncertainty-weighted TSDF volume} that reflects both geometric consistency and observation reliability across views.

\paragraph{TSDF-to-opacity mapping.}
To convert the truncated signed distance field $F(\mathbf{x}_j)$ into voxel opacity values, 
we define a differentiable mapping $\phi: F \mapsto \alpha$, 
which controls the opacity transition around the zero-level surface. 
We adopt a parametric sigmoid function
\begin{equation}
\phi(F) = 
\frac{1}{1 + \exp(-\varepsilon F)},
\end{equation}
where the slope $\varepsilon$ is analytically chosen such that 
a voxel located at $F = -0.1$ attains an opacity $\phi(F) = p_{\!0.1}$, 
typically $p_{\!0.1}=0.9$. 
This calibration guarantees a consistent surface thickness proportional to the truncation band.
The calibrated mapping provides an interpretable control over the opacity falloff 
and prevents over-thickening of surfaces.

\paragraph{Opacity normalization and pruning.}
The resulting opacity values are normalized to $[\alpha_{\min}, \alpha_{\max}]$ and assigned 
to all voxels in $V^{\mathrm{g}}$, producing the final representation $V^{\mathrm{f}}$. 
Voxels whose maximum opacity across their corner samples falls below a threshold $\tau_{\mathrm{prune}}$ 
are removed:
\begin{equation}
    \mathcal{M}_{\mathrm{prune}} = 
    \{\, v \in V^{\mathrm{f}} ~|~ 
    \max_{j \in \text{corners}(v)} \phi(F(\mathbf{x}_j)) < \tau_{\mathrm{prune}} \,\}.
\end{equation}
This pruning step discards low-confidence empty space while preserving regions 
with consistent, high-opacity support. The resulting fused octree $V^{\mathrm{f}}$ serves as a complete, continuous volumetric initialization for subsequent per-scene SVR optimization.


\subsection{Refined Depth for Geometry Supervision}\label{sec:depth_loss}

We optimize the geometry of the octree from Sec.~\ref{sec:voxel_init} with per-scene training. Unlike previous works~\cite{ligeosvr, chen2024pgsr} that regularize the geometry with an indirect, weighted unprojection loss, we adopt a local, per-pixel geometry-supervision loss that directly regularizes the geometry through depth refinement. We now detail the process below.

\subsubsection{Depth Refinement} For each training camera $C$ and pixel $p$, we form a compact set of refined depth candidates anchored at the rendered depth $\hat D_t^{(C)}(p)$, its two spatial neighbors, and small multiplicative perturbations: 
\begin{align*}
\mathcal{D}(p)
&= \big\{\hat D_t^{(C)}(p),\,
       \hat D_t^{(C)}(p+\mathbf{e}_x),\,
       \hat D_t^{(C)}(p+\mathbf{e}_y)\big\} \\
&\cup \big\{\hat D_t^{(C)}(p)(1+\epsilon_k)\big\}_{k=1}^{n_{\mathrm{rand}}}.
\end{align*}
with \(\epsilon_k \sim \mathrm{Unif}\!\big(-\tfrac{s}{2},\tfrac{s}{2}\big)\), \(s \ll 1\).

Each candidate is backprojected from $C$ and reprojected into $K$ nearest views $\mathcal{N}(C)$. We score each candidate by normalized cross-correlation (NCC) between fixed-size, bilinear-sampled patches and take the candidate with highest score as the refined depth $D_t^{\star}(p)$ of the pixel $p$. Then, a lightweight cross-view geometric check rejects the refined depths whose reprojection error exceeds a small threshold. The refinement produces a sparse, high-precision depth map $D_t^{\star}$.

\begin{table*}[t]
  \centering
  \begin{threeparttable}
  \scriptsize
  \setlength{\tabcolsep}{3.6pt}
  \renewcommand{\arraystretch}{1.15}
  \begin{tabularx}{\textwidth}{@{} l *{15}{Y} @{} G @{} Y c @{}}
  \toprule
  Method &
  24 & 37 & 40 & 55 & 63 & 65 & 69 & 83 & 97 & 105 & 106 & 110 & 114 & 118 & 122 & & Mean & Time \\
  \midrule
  \multicolumn{18}{l}{\textit{Implicit}} \\
  \midrule
  VolSDF {\cite{yariv2021volume}} &
  1.14 & 1.26 & 0.81 & 0.49 & 1.25 & 0.70 & 0.72 & 1.29 & 1.18 & 0.70 & 0.66 & 1.08 & 0.42 & 0.61 & 0.55 & & 0.86 & $>12$h \\
  NeuS {\cite{wang2021neus}} &
  1.00 & 1.37 & 0.93 & 0.43 & 1.10 & 0.65 & 0.57 & 1.48 & 1.09 & 0.83 & 0.52 & 1.20 & 0.35 & 0.49 & 0.54 & & 0.84 & $>12$h \\
  Neurangelo {\cite{li2023neuralangelo}} &
  \Y{0.37} & 0.72 & \Y{0.35} & \Y{0.35} & 0.87 & 0.54 & 0.53 & 1.29 & 0.97 & 0.73 & \Y{0.47} & 0.74 & \Y{0.32} & 0.41 & 0.43 & & 0.61 & $>128$h \\
  GeoNeuS {\cite{fu2022geo}} &
  0.38 & \Y{0.54} & \T{0.34} & 0.36 & 0.80 & \R{0.45} & \R{0.41} & \T{1.03} & \Y{0.84} & \R{0.55} & \T{0.46} & \T{0.47} & \R{0.29} & \Y{0.36} & 0.35 & & \Y{0.51} & $>12$h \\
  MonoSDF {\cite{yu2022monosdf}} &
  0.66 & 0.88 & 0.43 & 0.40 & 0.87 & 0.78 & 0.81 & 1.23 & 1.18 & 0.66 & 0.66 & 0.96 & 0.41 & 0.57 & 0.51 & & 0.73 & 6h \\
  \midrule
  \multicolumn{18}{l}{\textit{Explicit}} \\
  \midrule
  2DGS {\cite{huang20242d}} &
  0.48 & 0.91 & 0.39 & 0.39 & 1.01 & 0.83 & 0.81 & 1.36 & 1.27 & 0.76 & 0.70 & 1.40 & 0.40 & 0.76 & 0.52 & & 0.80 & 0.2h \\
  GOF {\cite{yu2024gaussian}} &
  0.50 & 0.82 & 0.37 & 0.37 & 1.12 & 0.74 & 0.73 & 1.18 & 1.29 & 0.68 & 0.77 & 0.90 & 0.42 & 0.66 & 0.49 & & 0.74 & 1h \\
  SVRaster {\cite{sun2025sparse}} &
  0.61 & 0.74 & 0.41 & 0.36 & 0.93 & 0.75 & 0.94 & 1.33 & 1.40 & 0.61 & 0.63 & 1.19 & 0.43 & 0.57 & 0.44 & & 0.76 & 0.1h/0.05h* \\
  GS2Mesh {\cite{wolf2024gs2mesh}} &
  0.59 & 0.79 & 0.70 & 0.38 & \Y{0.78} & 1.00 & 0.69 & 1.25 & 0.96 & \Y{0.59} & 0.50 & 0.68 & 0.37 & 0.50 & 0.46 & & 0.68 & 0.3h \\
  VCR-GauS {\cite{chen2024vcr}} &
  0.55 & 0.91 & 0.40 & 0.43 & 0.97 & 0.95 & 0.84 & 1.39 & 1.30 & 0.90 & 0.76 & 0.92 & 0.44 & 0.75 & 0.54 & & 0.80 & $\sim$1h \\
  MonoGSDF {\cite{li2024monogsdf}} &
  0.45 & 0.65 & 0.36 & 0.36 & 0.94 & 0.70 & 0.67 & 1.27 & 0.99 & 0.63 & 0.49 & 0.84 & 0.39 & 0.53 & 0.47 & & 0.65 & hrs \\
  PGSR {\cite{chen2024pgsr}} &
  \T{0.36} & 0.57 & 0.38 & \T{0.33} & \Y{0.78} & 0.58 & \Y{0.50} & \Y{1.08} & \T{0.63} & \Y{0.59} & \T{0.46} & \Y{0.54} & 0.30 & 0.38 & \Y{0.34} & & 0.52 & 0.5h \\
  GeoSVR {\cite{ligeosvr}}&
  \R{0.32} & \T{0.51} & \R{0.30} & \T{0.33} & \T{0.71} & \T{0.48} & \T{0.42} & \T{1.03} & \R{0.62} & \T{0.56} & \R{0.33} & \R{0.46} & \T{0.30} & \T{0.34} & \T{0.32} & & \T{0.47} & 0.8h/0.2h* \\
  \midrule
  \textbf{Ours} &
  \R{0.32} & \R{0.50} & \R{0.30} & \R{0.32} & \R{0.69} & \Y{0.50} & \T{0.42} & \R{1.00} & \R{0.62} & \R{0.55} & \R{0.33} & \R{0.46} & \T{0.30} & \R{0.32} & \R{0.31} & & \R{0.46} & 0.4h* \\
  \bottomrule
  \end{tabularx}
      \begin{tablenotes}[flushleft]
    \footnotesize
    \item (*) represents training time on RTX 5090.
  \end{tablenotes}
  \caption{Qualitative Comparison on the DTU Dataset~\cite{jensen2014large}. Each row represents a baseline method and each column represents a scene in the DTU dataset.}
  \label{tab:dtu-per-scan}
  \end{threeparttable}
\end{table*}

\subsubsection{Optimization Objectives} We supervise the geometry with an $\ell_1$ loss between the rendered depth and the refined target, $\mathcal{L}_{\mathrm{refD}}=|D_t-D_t^{\star}|$. The total objective is composed of the photometric loss $\mathcal{L}_{\mathrm{photo}}$ from SVRaster, the regularization term $\mathcal{L}_{\mathrm{reg}}$ from GeoSVR excluding the multi-view geometry loss, and our geometry-supervision loss $\mathcal{L}_{\mathrm{refD}}$:
\begin{equation}
    \mathcal{L} = \lambda_{\mathrm{photo}}\mathcal{L}_{\mathrm{photo}} + \lambda_{\mathrm{reg}}\mathcal{L}_{\mathrm{reg}} + \lambda_{\mathrm{refD}}\mathcal{L}_{\mathrm{refD}}
\end{equation}
In this work, we set the weights of $\lambda_{\mathrm{photo}} = 0.1$, $\lambda_{\mathrm{reg}} = 1.0$, and $\lambda_{\mathrm{refD}} = 0.05$, respectively.

Targets $D_t^{\star}$ are refreshed every $R$ iterations so supervision tracks the improving geometry without backpropagating through the discrete selection. With a good octree initialization from Sec.~\ref{sec:voxel_init}, the refined depth candidates concentrates near the true surface as the geometry improves while the local surfaces still remains sharp.


\section{Experiments}
\begin{figure}[t]
  \centering
  \includegraphics[width=\linewidth,trim=0 0 0 0,clip]{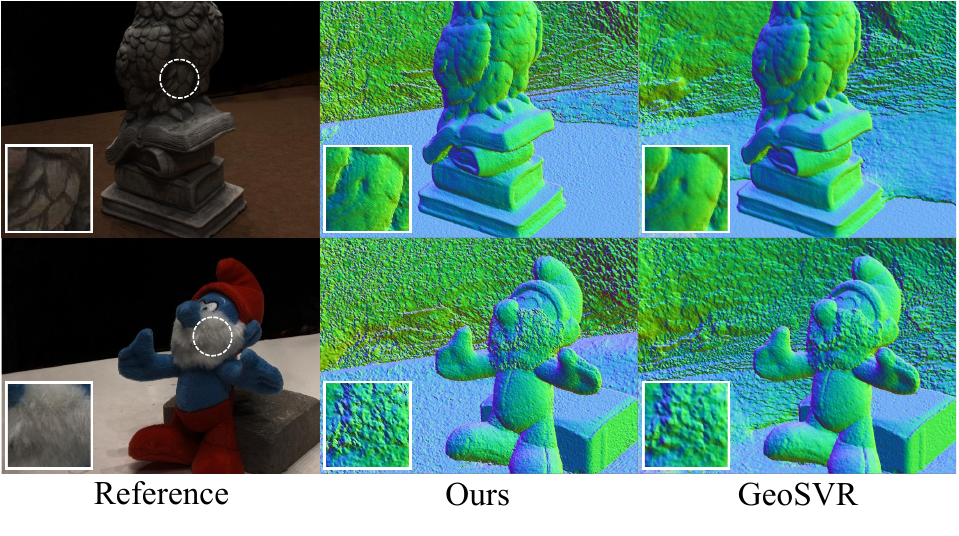}
  \caption{\textbf{Rendered Normals on DTU Dataset.} Our method preserves the local details while GeoSVR produces overly smoothed results.}
  \label{fig:vis_normal}
\end{figure}

\subsection{Implementation Details}
Our code is implemented with PyTorch and CUDA kernels, built atop the GeoSVR~\cite{ligeosvr} codebase.
Unless stated otherwise, each model is trained for $20\,000$ iterations with Adam~\cite{adam2014method}.
We use separate learning rates for different parameter groups: density $5{\times}10^{-2}$, zero-degree SH $1{\times}10^{-2}$, and higher-degree SH $2.5{\times}10^{-4}$.
Pseudo depth priors are generated by VGGT~\cite{wang2025vggt}, and DepthAnythingV2~\cite{yang2024depth} provides additional depth cues, following the same usage as in GeoSVR.
We use $7{\times}7$ patches for patch warping and depth refinement.
The voxel-dropout parameter is set to $\gamma{=}0.5$ on DTU.
Our octree setup follows SVRaster with the pruning interval increased to $2\,000$ iterations to encourage finer detail.
All experiments are run on a single NVIDIA RTX~5090 GPU.

\subsection{Surface Reconstruction}
We evaluate on DTU using the standard 15 scans.
Images are downsampled ${\times}2$, meshes are extracted with TSDF (voxel size $0.002$), and Chamfer distance (CD) is reported per scan and as the mean.
These settings match the common DTU protocol used in recent surface-reconstruction works.

\paragraph{Baselines and Timing.}
We compare to implicit methods including VolSDF~\cite{yariv2021volume}, NeuS~\cite{wang2021neus}, Neuralangelo~\cite{li2023neuralangelo}, Geo-NeuS~\cite{fu2022geo}, and MonoSDF~\cite{yu2022monosdf}, and explicit methods including 2DGS~\cite{huang20242d}, GOF~\cite{yu2024gaussian}, SVRaster~\cite{sun2025sparse}, GS2Mesh~\cite{wolf2024gs2mesh},
VCR-GauS~\cite{chen2024vcr}, MonoGSDF~\cite{li2024monogsdf}, PGSR~\cite{chen2024pgsr}, and GeoSVR~\cite{ligeosvr}.
For methods we did not re-run, we quote the numbers from the authors’ papers.
For SVRaster and GeoSVR we report both their published 3090 Ti times and our 5090 re-runs.

\begin{figure}[t]
  \centering
  \includegraphics[width=\linewidth,trim=0 0 0 0,clip]{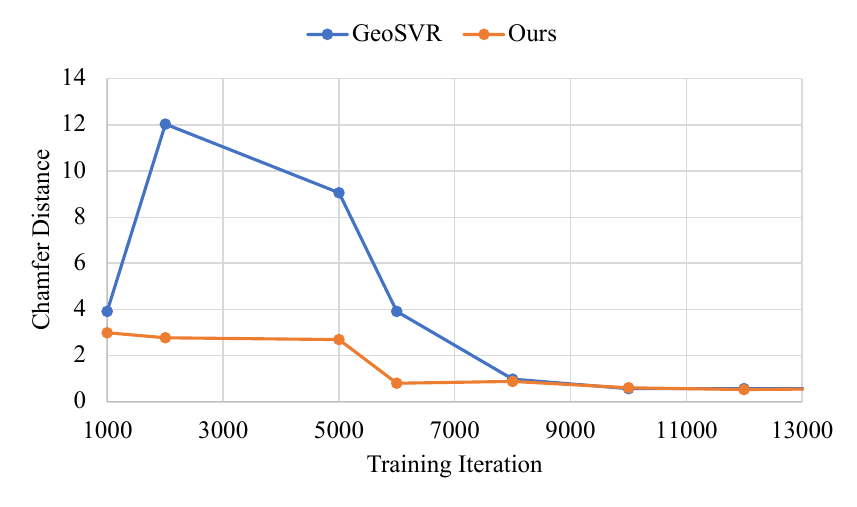}
  \caption{\textbf{Convergence on DTU Dataset.} Our method stabilizes training and accelerates convergence compared to the baseline. Note that, without specific geometry regularization, SVRaster~\cite{sun2025sparse} is not expected to outperform GeoSVR in terms of convergence.}
  \label{fig:convergence}
\end{figure}

\paragraph{Results and Discussion.}
Per-scan CD and training times are summarized in Tab.~\ref{tab:dtu-per-scan}. 
Our method attains the lowest mean CD on DTU while remaining efficient. 
3DGS-based approaches lack an explicit surface parameterization. Consequently, even when regularized with external geometric cues from pretrained models (e.g., GS2Mesh, VCR-GauS, MonoGSDF), their reconstructions still trail sparse-voxel methods (GeoSVR and ours). 
Sparse-voxel baselines (SVRaster, GeoSVR) start from uniform grids. Without strong initialization, geometric regularizers tend to bias the solution toward overly smooth surfaces (see Fig.~\ref{fig:vis_normal}). 
The voxel initialization with level of detail and refined depth for geometry supervision aids the SVO in preserving crisp, well-defined surfaces without over-smoothing, yielding consistent gains across scans. 

We discuss the benefits of our method in convergence speed on DTU, as shown in Fig.~\ref{fig:convergence}. We report the average CD over the first ten scenes. Compared to the baseline, our initialization provides accurate starting point for per-scene optimization, stabilizes the training process, and allows the optimization to converge in fewer iterations than prior sparse-voxel rasterization baselines.
In contrast, the training process of sparse-voxel baseline that starts with a uniform grid is highly unstable at the beginning, which leads to large geometry fluctuations as voxels are pruned and subdivided.

\begin{table}[t]
    \centering
    \footnotesize
    \setlength{\tabcolsep}{4pt} 
    \begin{tabular*}{\columnwidth}{@{\extracolsep{\fill}} l c @{}}
        \toprule
        Method & CD ($\downarrow$)\\
        \midrule
        GeoSVR (base) & 0.47 \\
        GeoSVR + voxel initialization & 0.47 \\
        GeoSVR + geometry supervision & 0.48 \\
        GeoSVR + geometry supervision + voxel  initialization (ours) & 0.46 \\
        \bottomrule
    \end{tabular*}
    \caption{\textbf{Ablation Study} on the DTU Dataset. We adopt GeoSVR as our baseline approach as it reconstructs surfaces with better geometry than SVRaster.}
    \label{tab:placeholder}
\end{table}

\subsection{Ablation Studies}
\label{sec:ablation}

We study the contribution of Voxel Initialization with Level of Details (Sec.~\ref{sec:voxel_init}) and Refined Depth for Geometry Superivision (Sec.~\ref{sec:depth_loss}) with four variants on DTU, keeping preprocessing, training schedule, and loss weights fixed. Table~\ref{tab:placeholder} reports mean Chamfer distance (CD) for all variants.

Adding \textbf{Refined Depth for Geometry Supervision} (GeoSVR + geometry supervision) without Voxel Initialization with Level of Details does not improve the GeoSVR baseline. This is expected since our refinement constructs depth candidates locally around the current prediction for each pixel. When the starting surface is far away from the actual surface, the depth candidates cluster near an incorrect depth, NCC becomes less discriminative, and the cross-view geometric check prunes many pixels. The resulting pseudo-labels are sparse or noisy, so the supervision signal is too weak to correct the geometry.

Adding \textbf{Voxel Initialization with Level of Details} (GeoSVR + voxel initialization) without the Refined Depth for Geometry Supervision keeps the mean CD unchanged. Although our voxel initialization algorithm assigns voxel levels by pixel footprint and yields a better-aligned octree, the weighting of the \emph{indirect} supervision of the reprojection loss in GeoSVR is ineffective in refining the geometry. When alignment and depth cues are still imperfect, the masking of GeoSVR reprojection loss drops many pixels and the weighting suppresses gradients where errors are large. Gradients therefore remain weak around thin structures and early misalignments. Swapping the initialization without an effective supervision in training does not translate into lower CD.

Enabling \textbf{both Voxel Initialization with Level of Details and Refined Depth for Geometry Supervision} (ours) improves the mean CD. With our Voxel Initialization with Level of Details, the refined depth candidates are sampled around near-correct geometry. The patch NCC scores in the depth refinement becomes more discriminative, geometric verification retains substantially more pixels, and the resulting refined depths provide \emph{direct}, dense, voxel-aligned supervision. The combination of a good initialization and direct depth supervision yields accurate pseudo-labels and drives consistent surface improvement.

In our ablations, both the GeoSVR reprojection loss and our refinement can reject many pixels when geometry is far from the truth, but they differ fundamentally. The GeoSVR term is an \emph{indirect} image-space penalty with exponential down-weighting and masking. When errors are large, weights vanish and gradients point in oblique directions that only weakly correct depth. Our refinement converts multi-view evidence into \emph{direct} per-ray metric depth targets; whenever a label is not masked, the $\ell_1$ signal is strong, well-conditioned, and aligned with the depth axis. After our voxel initialization brings the surface into the right basin, our refined depth candidates include near-true hypotheses and improves geometry.

\section{Conclusion}

In this work, we present a sparse-voxel framework that bridges voxel representation with scene structure for high-quality, topology-friendly surface reconstruction. We pair a principled octree initialization that allocates capacity only where multi-view evidence supports it with a direct geometry supervision that updates per-pixel depths, yielding reconstructions that preserve fine detail, maintain high completeness, and converge in fewer iterations than prior sparse-voxel approaches. On standard benchmarks, this design improves geometric accuracy over strong baselines while remaining efficient, offering a practical recipe for radiance-field surfaces that are easy to optimize and export as meshes. Looking ahead, we aim to extend this principle to challenging illumination and texture-poor regions via uncertainty-aware view selection and multi-scale cues, to large-scale scenes through scalable octree expansion, and to richer priors (e.g., normals or semantics) that further stabilize early training and enhance robustness.
{
    \small
    \bibliographystyle{ieeenat_fullname}
    \bibliography{main}
}
\setcounter{figure}{0}
\setcounter{table}{0}
\renewcommand{\thefigure}{A\arabic{figure}}
\renewcommand{\thetable}{A\arabic{table}}

\clearpage
\setcounter{page}{1}
\maketitlesupplementary
\section*{A. Details on TSDF-to-Opacity Mapping}
\label{sec:opacity}

\begin{figure}
    \centering
\subfloat[]{
    \includegraphics[width=0.95\linewidth]{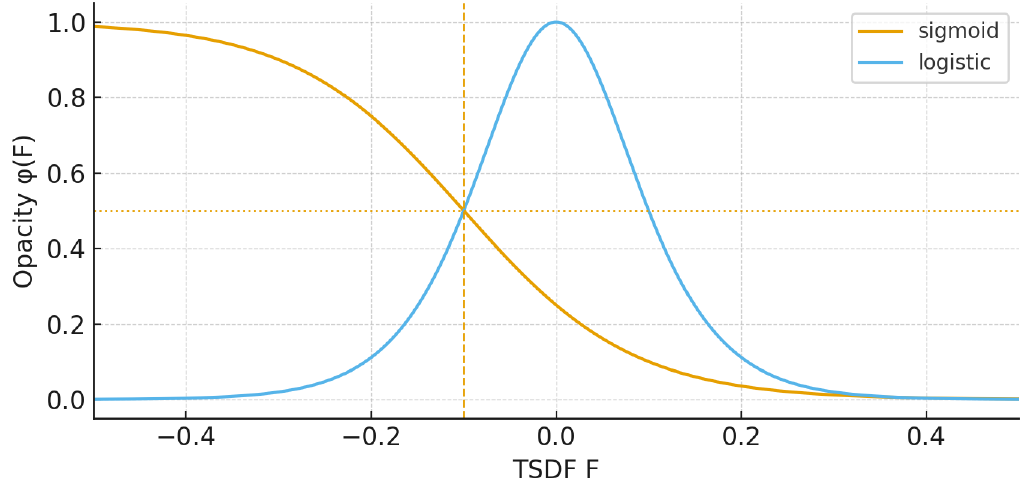}
    \label{fig:mapping}
}\\
\subfloat[]{
    \includegraphics[width=0.95\linewidth]{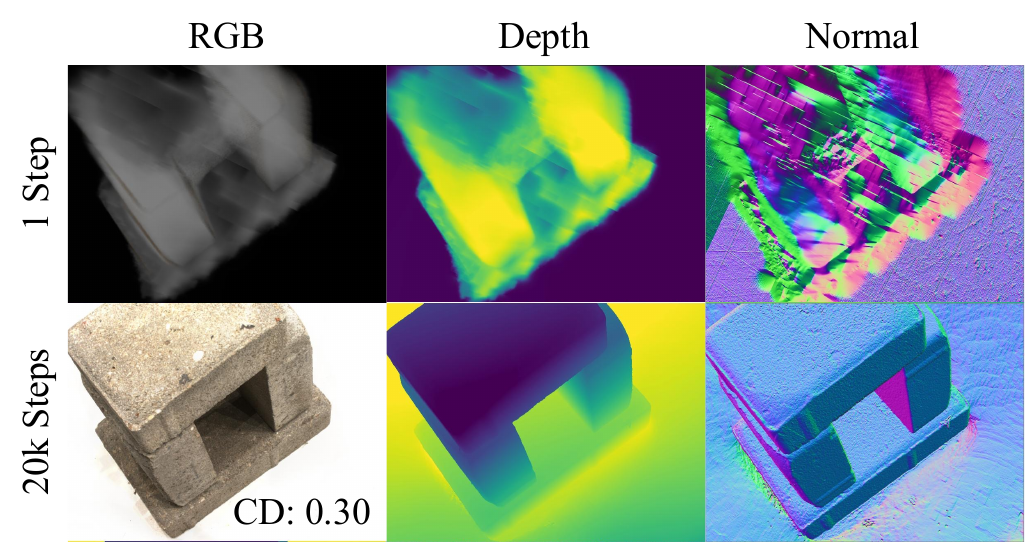}
    \label{fig:sigmoid}
}\\
\subfloat[]{
    \includegraphics[width=0.95\linewidth]{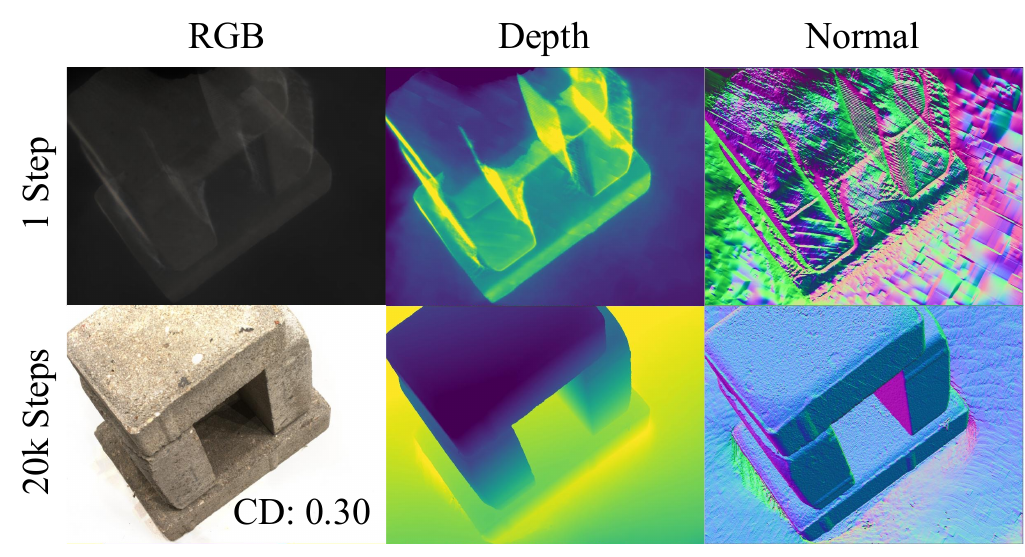}
    \label{fig:bell}
}
    \caption{Visualization of TSDF-to-opacity mapping. (a) The sigmoid-like function and the logistic density function. Both with $p=0.5$, which means the opacity before scaling is $0.5$ at $\mathrm{TSDF}=-0.1$. (b) Visualization of scene-specific reconstruction with sigmoid-like TSDF-to-opacity mapping in the voxel initialization. (c) Visualization of scene-specific reconstruction with logistic TSDF-to-opacity mapping in the voxel initialization. Both initialization yield the same geometric quality, which results in same CD.}
    \label{fig:mapping_and_vis}
\end{figure}

\paragraph{Setup.}
Let $F(\mathbf{x})$ be a truncated signed distance field (TSDF) with $F<0$ inside the surface and $F>0$ outside.
We construct a differentiable mapping $\phi: F \mapsto \alpha \in [0,1]$ that controls how opacity rises around the zero level set.
We demonstrate a sigmoid-like mapping and a logistic mapping (shown in Fig.~\ref{fig:mapping_and_vis})
We denote by $a>0$ a small offset (e.g., $a=0.1$ in normalized TSDF units).

\subsection*{A.1 Sigmoid-Like Mapping}
We map the voxel grids inside the surface to high opacity and grids outside the surface to low opacity with a sigmoid-like function parameterized by transition $\beta$:
\begin{equation}
\phi_{\mathrm{sig}}(F;\,\beta)\;=\;\sigma\!\left(-\frac{F}{\beta}\right)\;=\;\frac{1}{1+\exp\!\bigl(\frac{F}{\beta}\bigr)}.
\label{eq:sigmoid}
\end{equation}
\textbf{Point calibration (equal-thickness):} choose $\beta$ so that $\phi_{\mathrm{sig}}(-a)=p$ (e.g., $p=0.9$):
\begin{equation}
p \;=\; \frac{1}{1+\exp(a/\beta)}
\;\Longrightarrow\;
\beta \;=\; -\,\frac{a}{\ln\!\bigl(\tfrac{1}{p}-1\bigr)}.
\label{eq:sig-cal}
\end{equation}
$\beta$ sets the transition sharpness so that a voxel just inside the surface ($F=-a$) reaches a prescribed opacity $p$,
yielding a consistent “thickness” across scenes.

\subsection*{A.2 Logistic Density Mapping}
Motivated by NeuS~\cite{wang2021neus}, we also consider a \emph{bell-shaped} opacity concentrated near $F{=}0$:
\begin{equation}
\phi_{\mathrm{bell}}(F;\,s)
\;=\;
\frac{4\,e^{-sF}}{ \bigl(1+e^{-sF}\bigr)^2 } \;\in [0,1],
\label{eq:bell}
\end{equation}
i.e., the logistic \emph{pdf} normalized so its peak is $1$ at $F{=}0$ (a thin ``opacity shell''). We enforce $\phi_{\mathrm{bell}}(-a)=b$ (e.g., $b{=}0.5$)
to match the inside thickness at $F=-a$.
Solving $4u/(1+u)^2=b$ for $u=e^{-s(-a)}=e^{sa}$ gives
$u=({2-b\pm 2\sqrt{1-b}})/b$ and $s=(\ln u)/a$ (pick the $u>1$ root for $s>0$).

\subsection*{A.3 From $\alpha$ to Stored Grid Values and Pruning}
The mapped opacity of each grid point is rescaled to $[\texttt{min\_opacity},\,\texttt{max\_opacity}]$ and voxels with opacity below a predefined threshold are pruned.
Comparing the two mapping functions, $\phi_{\mathrm{bell}}$ concentrates opacity more tightly than $\phi_{\mathrm{sig}}$ and typically produces \emph{fewer} above-threshold voxels (thinner band),
leading to a \emph{sparser} octree around the surface at the same pruning threshold.

\begin{table*}[t]
  \centering
  \begin{threeparttable}
  \scriptsize
  \setlength{\tabcolsep}{3.6pt}
  \renewcommand{\arraystretch}{1.15}
  \begin{tabularx}{\textwidth}{@{} l *{15}{Y} @{} G @{} Y @{}}
  \toprule
  Foundation Model &
  24 & 37 & 40 & 55 & 63 & 65 & 69 & 83 & 97 & 105 & 106 & 110 & 114 & 118 & 122 & & Mean \\
  \midrule
  \textbf{VGGT} &
  {0.32} & {0.50} & {0.30} & {0.32} & {0.69} & {0.50} & {0.42} & {1.00} & {0.62} & {0.55} & {0.33} & {0.46} & {0.30} & {0.32} & {0.31} & & {0.46} \\
  \midrule
  \textbf{Depth Anything 3} &
  {0.32} & {0.49} & {0.29} & {0.32} & {0.71} & {0.50} & {0.43} & {1.00} & {0.62} & {0.55} & {0.32} & {0.46} & {0.30} & {0.32} & {0.31} & & {0.46} \\
  \bottomrule
  \end{tabularx}
  \caption{Quantitative Comparison on Different Foundation Model Priors. Each row represents Chamfer Distance of reconstruction result from our method with prior from the corresponding 3D foundation model. Each column represents a scene in the DTU dataset. The results show that our method is foundation model-agnostic with similar performance across different foundation model priors.}
  \label{tab:fms_quant}
  \end{threeparttable}
\end{table*}

\section*{B. Alternative 3D Foundation Models}
\label{sec:sup:fm-agnostic}

Our pipeline is agnostic to the choice of 3D foundation model. In practice, we require only three cues per image: calibrated cameras $C_i=(\mathbf{K}_i,\mathbf{R}_i,\mathbf{t}_i)$, dense (or reasonably dense) depth $\tilde{D}_i$, and an optional confidence map $\tilde{\kappa}_i$. Any foundation model that supplies these can be used via a thin adapter: we align predictions to our reconstruction frame with a global similarity transform (scale $s$ and rigid $(\mathbf{R},\mathbf{t})$), rescale depths $D_i=s\,\tilde{D}_i$, and pass $(C_i,D_i,\kappa_i)$ to our Voxel Initialization with Level of Details (Sec.~\ref{sec:voxel_init}). No point-map or track outputs are required. When a confidence map is unavailable, we use multi-view depth consistency as a proxy for voxel initialization.

\begin{figure}[t]
\centering
\includegraphics[width=\linewidth]{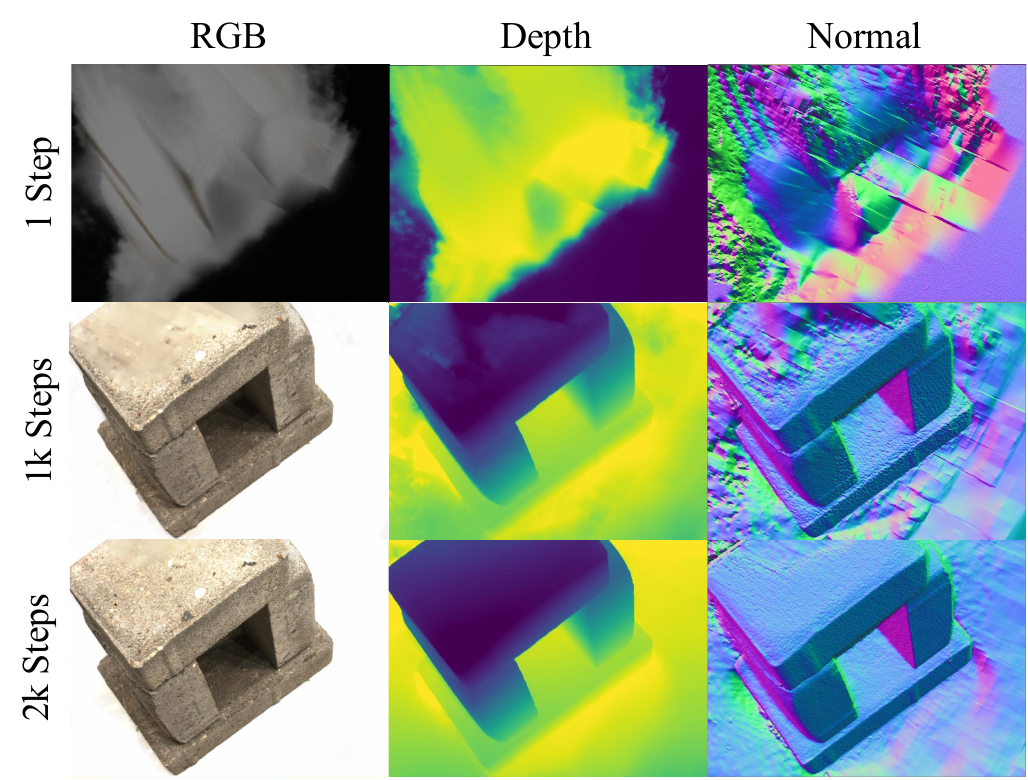}
\caption{\textbf{Replacing the 3D foundation model.}
Our method runs unchanged when swapping VGGT for another 3D foundation model (here, Depth Anything 3).
Results remain crisp at early iterations and converge rapidly, illustrating that the pipeline is foundation-model agnostic.}
\label{fig:da3}
\end{figure}

To demonstrate this foundation model-agnostic property, beyond VGGT we also ran the full pipeline using a different 3D foundation model Depth Anything 3~\cite{lin2025depth3recoveringvisual}. Table~\ref{tab:fms_quant} compares the geometry accuracy in Chamfer Distance on DTU dataset and shows that different foundation model priors achieve the same quantitative results. The qualitative results in Fig.~\ref{fig:da3} show that our initialization and refinement behave consistently—preserving early sharpness and converging quickly—without any changes to losses, schedules, or octree settings.

\section*{C. Mesh Visualization}
\begin{figure*}
    \centering
    \includegraphics[width=\textwidth,trim=10 0 10 0,clip]{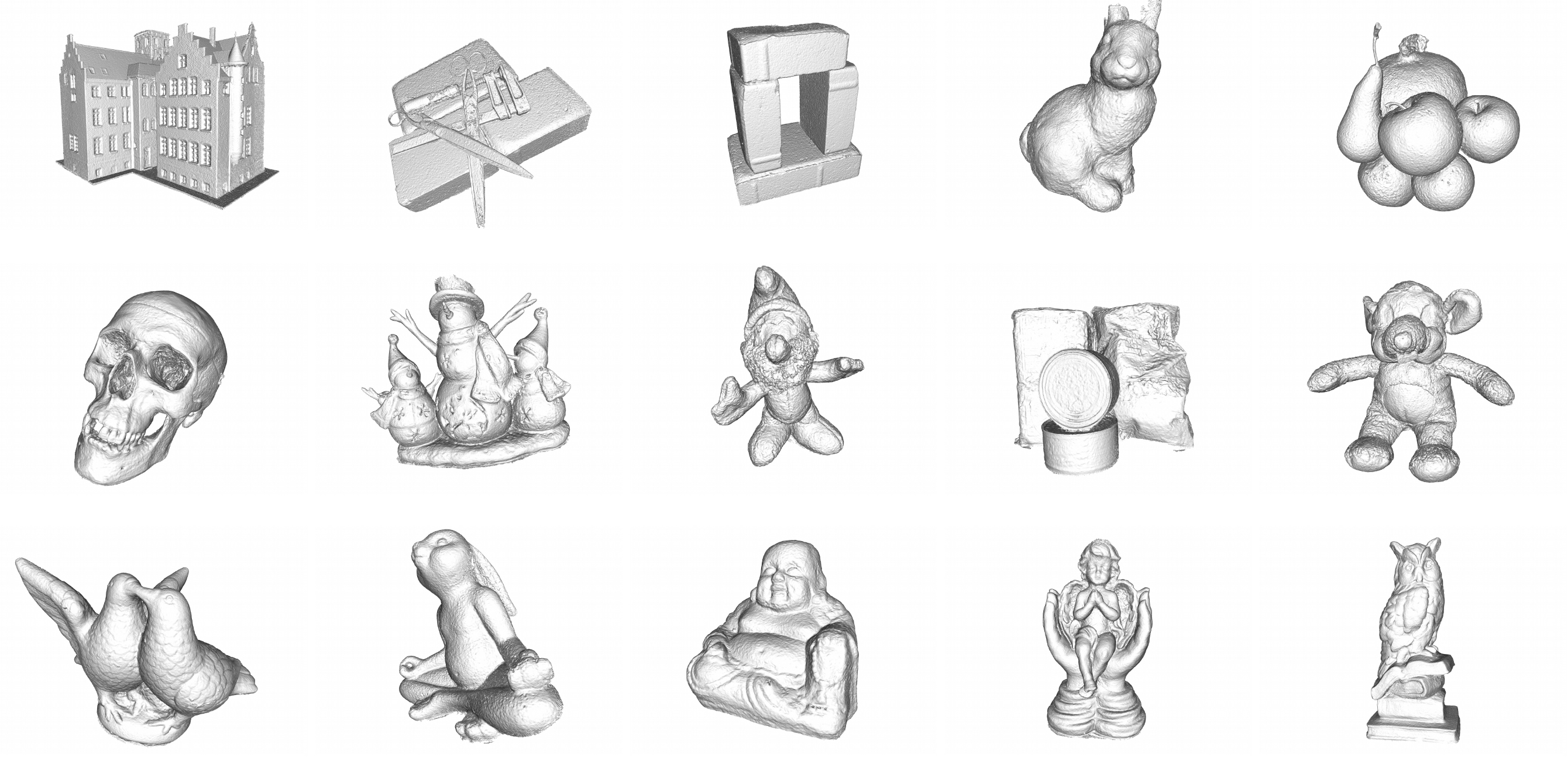}
    \caption{Qualitative results of the reconstructed mesh.}
    \label{fig:dtu_mesh}
\end{figure*}

We reconstruct meshes from our trained sparse voxel octrees using the official TSDF-based mesh extraction method provided by the DTU benchmark. The visualization in Fig.~\ref{fig:dtu_mesh} shows that the resulting meshes are smooth while preserving fine detail.
\label{sec:sup:fm-agnostic}

\end{document}